\documentclass[final, 11pt, authoryear,twocolumn]{elsarticle}
\usepackage{amssymb}
\usepackage{pifont}
\usepackage{graphicx}
\usepackage{subfig}
\usepackage{color}
\usepackage{natbib}
\usepackage{amsmath}
\usepackage{algorithm}
\usepackage{algpseudocode}
\usepackage{verbatim}

\usepackage{geometry}
\begin{document}
\begin{frontmatter}
\title{One-shot domain adaptation in multiple sclerosis lesion segmentation using convolutional neural networks}
\author[label1]{Sergi Valverde \corref{corr1}}
\author[label1,label2]{Mostafa Salem}
\author[label1]{Mariano~Cabezas}
\author[label3]{Deborah Pareto}
\author[label4]{Joan~C.~Vilanova}
\author[label5]{Llu\'is Rami\'o-Torrent\`a}
\author[label3]{\`Alex~Rovira}
\author[label1]{Joaquim~Salvi}
\author[label1]{Arnau~Oliver}
\author[label1]{Xavier~Llad{\'o}}

\address[label1]{Research institute of Computer Vision and Robotics, University of Girona, Spain}
\address[label2]{Computer Science Department, Faculty of Computers and Information, Assiut University, Egypt}
\address[label3]{Magnetic Resonance Unit, Dept of Radiology, Vall d'Hebron University Hospital, Spain}
\address[label4]{Girona Magnetic Resonance Center, Spain}
\address[label5]{Multiple Sclerosis and Neuroimmunology Unit, Dr. Josep Trueta University Hospital, Spain}

\cortext[corr1]{Corresponding author. S. Valverde, Ed. P-IV, Campus Montilivi, University of Girona, 17003 Girona (Spain).
e-mail: svalverde@eia.udg.edu. Phone: +34 972 418878; Fax: +34 972 418976.}

\begin{abstract}
In recent years, several convolutional neural network (CNN) methods have been proposed for the automated white matter lesion segmentation of multiple sclerosis (MS) patient images, due to their superior performance compared with those of other state-of-the-art methods. However, the accuracies of CNN methods tend to decrease significantly when evaluated on different image domains compared with those used for training, which demonstrates the lack of adaptability of CNNs to unseen imaging data. In this study, we analyzed the effect of intensity domain adaptation on our recently proposed CNN-based MS lesion segmentation method. Given a source model trained on two public MS datasets, we investigated the transferability of the CNN model when applied to other MRI scanners and protocols, evaluating the minimum number of annotated images needed from the new domain and the minimum number of layers needed to re-train to obtain comparable accuracy. Our analysis comprised MS patient data from both a clinical center and the public ISBI2015 challenge database, which permitted us to compare the domain adaptation capability of our model to that of other state-of-the-art methods. In both datasets, our results showed the effectiveness of the proposed model in adapting previously acquired knowledge to new image domains, even when a reduced number of training samples was available in the target dataset. For the ISBI2015 challenge, our one-shot domain adaptation model trained using only a single image showed a performance similar to that of other CNN methods that were fully trained using the entire available training set, yielding a comparable human expert rater performance. We believe that our experiments will encourage the MS community to incorporate its use in different clinical settings with reduced amounts of annotated data. This approach could be meaningful not only in terms of the accuracy in delineating MS lesions but also in the related reductions in time and economic costs derived from manual lesion labeling.
\end{abstract}

\end{frontmatter}

\newpage
\section{Introduction}
\label{sec:introduction}


Currently, magnetic resonance imaging (MRI) is extensively used in the diagnosis and monitoring of multiple sclerosis (MS), due to the sensitivity of structural MRI disseminating focal white matter (WM) lesions in time and space \citep{Rovira2015}. With different modifications of MRI criteria over time, the presence of new lesions on  MRI scans is considered a prognostic and predictive biomarker for the disease \citep{Filippi2016}. Although visual lesion inspection is feasible in practice, this task is time-consuming, prone to manual errors and variable for different expert raters, which has lead to the development of a wide number of automated strategies in recent years \citep{Llado2012}.

Although there is a  wide range of methods proposed, convolutional neural network (CNN) strategies are being increasingly introduced. In contrast to previously supervised learning methods, CNNs do not require manual feature engineering or prior guidance, which along with the increase in computing power makes them a very interesting alternative for automated lesion segmentation,  as seen by their top ranking performance on all of the international MS lesion challenges \citep{Styner2008, Commowick2016, Carass2017}. The proposed network architectures and training pipelines include three-dimensional (3D) encoder networks with shortcut connections \citep{Brosch2016}, multi-view image architectures \citep{Greenspan2016}, cascaded 3D pipelines \citep{Valverde2017}, multi-dimensional recurrent gated units \citep{Andermatt2017a} and fully convolutional architectures \citep{Roy2018, Hashemi2018}.

However, CNN architectures applied in MRI tend to not generalize well on unseen image domains, which is mostly due to variations in image acquisition, MRI scanner, contrast, noise level or resolution between image datasets. As a result,  manual expert labeling must be performed on the new image domain, which is very-time consuming and not always possible. In this aspect, only a few papers have analyzed the CNN domain adaptation problem on brain MRI. Recently, \citet{Kamnitsas2017} proposed an unsupervised domain adaptation CNN model for the segmentation of traumatic brain injuries, where adversarial training was applied to adapt two related image domains with distinct types of image sequences. Similarly, \citet{Ghafoorian2017b} investigated the transferability of the acquired knowledge of a CNN model that was initially trained for WM hyper-intensity segmentation on legacy low-resolution data when applied to new data from the same scanner but with higher image resolution, showing the minimum amount of supervision required in terms of high-resolution training samples and re-trained network layers. Nevertheless, in both studies neither the experiments nor the segmentation tasks were focused between completely unrelated MS image domains in terms of the image acquisition (scanner), resolution and contrast, which can be very interesting in evaluating the usability of these CNN models in different clinical scenarios.

{In this paper, we analyzed the effectiveness of supervised image domain adaptation between completely unrelated MS databases. To do so, we first trained a slightly modified version of our already proposed cascaded architecture \citep{Valverde2017} entirely using two public MS databases from the Medical Image Computing and Computer Assisted Intervention (MICCAI) society, MICCAI2008 \citep{Styner2008} and MICCAI2016 \citep{Commowick2016}, which was considered the \textit{source} model. Then, we analyzed the transferring knowledge capability of this model by evaluating its performance on a set of completely unseen images from other \textit{target} image domains, partly re-training a different number of layers or no layers. We extended our analysis to investigate the minimum number of unseen images and re-trained layers needed to obtain a similar performance on the domain adapted model, even in one-shot domain scenarios in which only a single training image was available on the target domain. Our evaluation included a clinical dataset and public MS data from the International Symposium on Biomedical Imaging (ISBI) 2015 MS challenge \citep{Carass2017}, comparing the performance of the domain-adapted CNN model with those of the same model fully trained on the target domain and other state-of-the-art methods. To promote the reproducibility and usability of our research, the proposed domain adaptation methodology is available as part of our \textit{nicMSlesions} MS lesion software, which can be downloaded freely from our research website\footnote{\texttt{http://github.com/NIC-VICOROB/nicmslesions}}.

\section{Materials and methods}
\label{sec:methods}

\subsection{CNN architecture}

\label{subsec:cnn_method}

The CNN MS lesion model follows our recently proposed framework for MS lesion segmentation \citep{Valverde2017}. Within this framework, a cascade of two identical CNNs are optimized, where the first network is trained to be more sensitive to revealing possible candidate lesion voxels, while the second network is trained to reduce the number of false positive outcomes. For a complete description of the details and motivations for the proposed architecture, please refer to the original publication.

\begin{figure*}[tp]
  \begin{center}
      \includegraphics[width=0.7\textwidth]{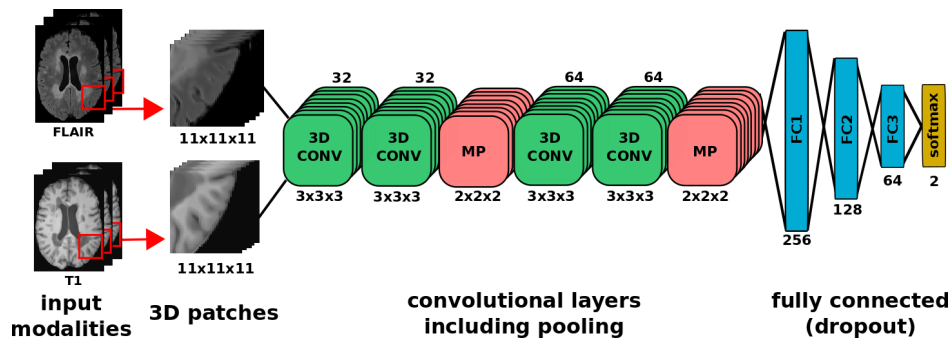}
  \end{center}
  \caption{Eleven-layer CNN model architecture trained using multi-sequence 3D image patches (FLAIR and T1-w) that are $11 \times 11 \times 11$ in size. Compared to the original implementation by \citet{Valverde2017}, we double the number of layers on each convolutional stack and add two additional fully connected layers of sizes 128 and 64, before the softmax layer.}
    \label{CNN_pipeline}
\end{figure*}

The architecture by \citet{Valverde2017} was composed of two stacks of convolution and max-pooling layers with 32 and 64 filters, respectively. The convolutional layers were followed by a fully connected (FC) layer of 256 in size and a softmax FC layer, summing $\sim$200K parameters. Here, to accommodate more expressive features that arise from the baseline training, we propose to double the number of layers on each convolutional stack (see Figure \ref{CNN_pipeline}). Additionally, we also stack two additional FC layers of size 128 and 64, to increase the number of potentially retrained classification layers used to adapt the image domains. The resulting CNN architecture consists of $\sim 470K$ network parameters.

The CNN training and inference procedures are identical to those proposed by \citet{Valverde2017}. Briefly, training is performed following a two-step approach: first, a CNN model is trained using a balanced set of multi-channel FLAIR and T1-w 3D $11\times 11 \times 11$ patches extracted from all of the available lesion voxels and a random selection of normal appearing tissue voxels. Then, the error of the first CNN model is computed by performing inferences on the same training set. Finally, the second model is trained using again a balanced set of voxels composed of all of the lesion voxels and a random selection of the misclassified lesion voxels on the previous model. Afterward, inferencing on the unseen images is performed by evaluating all of the input voxels using the first trained CNN, which discards all of the voxels with a low probability of being lesion. The remaining voxels are re-evaluated using the second CNN, obtaining a lesion probabilistic lesion mask. Final binary output masks are computed by linear thresholding of probabilities  $\ge t_{bin}$ and a posterior filtering of the resulting binary regions with a lesion size below $l_{min}$.

\subsection{Initial training}
\label{subsec:initial_training}

The proposed CNN architecture was first fully trained using 35 images from the two publicly available MS lesion segmentation datasets of the MICCAI society. Both the MICCAI2008 \citep{Styner2008} and MICCAI2016 \citep{Commowick2016} are currently used as benchmarks to compare the accuracy of novel MS lesion segmentation pipelines. Please note that for each individual challenge, the proposed network architecture performed in the top rank (see \citet{Valverde2017} for the final ranking and comparison with other state-of-the-art methods).

\subsubsection{MICCAI 2008 dataset}
The MICCAI 2008 MS lesion segmentation challenge was composed of 20 training scans from research subjects, which were acquired at Children's Hospital Boston (CHB, 3T Siemens) and University of North Carolina (UNC, 3T Siemens Alegra). For each subject, the original T1-w, T2-w and FLAIR image modalities were provided with an isotropic resolution of $0.5 \times 0.5 \times 0.5~mm^3$. The provided FLAIR and T2-w image modalities were already rigidly co-registered to the T1-w space. All of the subjects were provided with manual expert annotations of WM lesions from a CHB and UNC expert rater. As pointed out by \citet{Styner2008}, the UNC manual annotations were adapted to closely match those from CHB, and thus, only the CHB annotations were used.

As a previous step, we skull-stripped both the T1-w and FLAIR images using the Brain Extraction Tool (BET) \citep{Smith2002} and intensity normalized using N3 \citep{Sled1998}. All of the training images were then resampled to ($1 \times 1 \times 1~mm$) using the FSL-FLIRT utility \citep{Greve2009}.

\subsubsection{MICCAI 2016 dataset}

The MICCAI 2016 MS lesion segmentation challenge was composed of 15 training scans acquired in different image domains: 5 scans (Philips Ingenia 3T), 5 scans (Siemens Aera 1.5T) and 5 scans (Siemens Verio 3T). For each subject, 3D T1-w MPRAGE, 3D FLAIR, 3D T1-w gadolinium enhanced and 2D T2-w/DP images were provided, presenting different image resolutions for each image domain (see the organizer's website for the exact details of the acquisition parameter and image resolutions\footnote{\texttt{https://portal.fli-iam.irisa.fr/msseg-\\challenge/overview}}). Manual lesion annotations for each training subject were provided as a consensus mask among 7 different human raters.

Pre-processed images were already provided. The pre-processing pipeline consisted of a denoising step with the NL-means algorithm \citep{Coupe2008} and a rigid registration~\citep{Commowick2012} of all of the modalities against the FLAIR image. Then, each of the modalities were skull-stripped using the volBrain platform~\citep{Manjon2016} and intensity corrected using the N4 algorithm~\citep{Tustison2010}. Finally, all of the training images were resampled to the same voxel space ($1 \times 1 \times 1~mm$) using the FSL-FLIRT utility \citep{Greve2009}.

\subsubsection{Experiment details}
\label{subsec:source_details}

All of the training images were first normalized with a zero mean and a standard deviation of one. The normalized images were used to build a set of 1200000 training patches, where 25\% was selected for validation and the others were used to optimize the network's weights. We trained each of the two networks for 400 epochs with an early stopping of 50 for each network. The parametric rectified linear activation function (PReLU) \citep{He2015} was applied to all layers. The convolutional layers were regularized using batch normalization \citep{Ioffe2015}, while dropout \citep{Srivastava2014} was applied to each of the FCs with ($p=0.5$). Network optimization was performed using the adaptive learning rate method (ADADELTA) \citep{Zeiler2012} with a batch size of 128 and categorical cross-entropy as the loss cost. The post-processing parameters $\ge t_{bin}$ and $l_{min}$ were set to 0.5 and 10, respectively.

\subsection{Supervised domain adaptation}
\label{subsec:domain_adaptation}

Although the convolutional layers can encode domain independent valid image features that describe the location, shape and lesion contrast, these features are then propagated through the FC layers, which learn to classify the lesion voxels based on the training data. However, this process is inherently dependent on the training domain characteristics, such as the intensity ratio between the lesion and the normal appearing tissue, which enables the FC layers to learn to optimize the best correlation between the extracted convolutional layers and the manual labels.

\begin{figure}[tp]
  \vspace{-0.5cm}
  \begin{center}
      \includegraphics[width=0.45\textwidth]{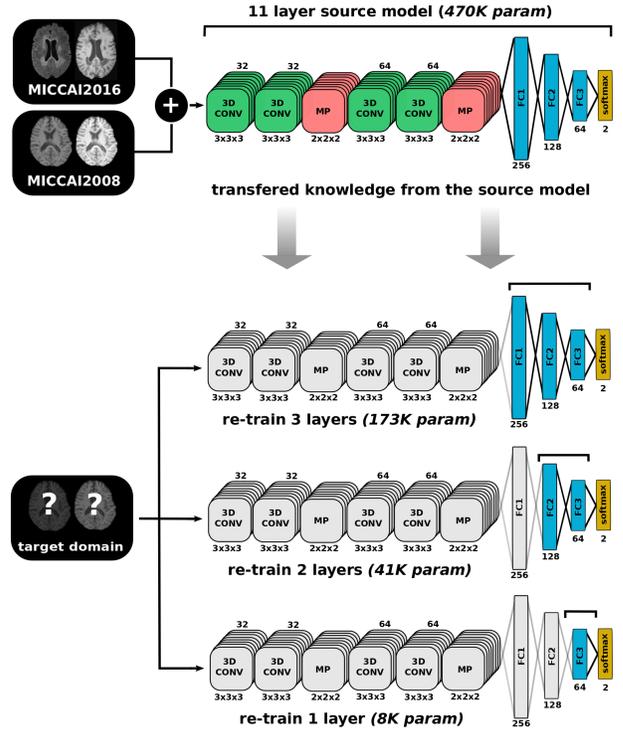}
  \end{center}
  \caption{Supervised intensity domain adaptation framework. From the 11 layer CNN source model trained on two public MS datasets (see Subsection \ref{subsec:initial_training}), we transfer the model knowledge to an unseen target image domain.  Domain adaptation is performed via 3 possible configurations by retraining the first FC layer, two FC layers or all FC layers using images and labels from the target intensity domain. In all of the configurations, the layers that are not re-trained are depicted in gray.}
    \label{domain_adaptation}
  \end{figure}

However, the encoded knowledge already present in the source model can be effectively used to adapt it to an unseen target intensity domain because convolutional layers contain  related features that can be transferred to unseen data while only re-training the FC layers (see Figure \ref{domain_adaptation}). In our experiments, domain adaptation is performed by retraining all or some of the source FC layers using images from the target domain. Table \ref{table:parameters} shows the number of network parameters used in each of the proposed configurations. As a result of reusing part of the implicit knowledge trained on the source model, the number of weights to optimize on the target model is significantly lower, which permits us to train the model with a reduced number of training images without over-fitting the model.

\subsection{Implementation}
All of the experiments were run on a GNU/Linux machine box running Ubuntu 16.04, with 32GB of RAM memory. CNN training was conducted on a single NVIDIA TITAN-X GPU (NVIDIA Corp, United States) with 12GB of RAM memory. All of the procedures were implemented in the Python language\footnote{\texttt{https://www.python.org/}}, using the Keras\footnote{\texttt{https://keras.io}} and Theano\footnote{\texttt{https://deeplearning.net/software/theano/}} \citep{Bergstra2011} libraries. The proposed method was integrated as part of our MS lesion segmentation software \textit{nicMSlesions}, which is available for downloading at our research website\textsuperscript{1}. 

\begin{table}[tp]
  \small
  \begin{center}
    \caption{Training parameters on each of the CNN models used. When training the source model (see Subsection \ref{subsec:initial_training}), all of the network layers are optimized from scratch. On the target models, only the last FC layer (FC3), last two FC layers (F2 + FC3) or all FC layers (FC1 + FC2 + FC3) are optimized, which significantly reduces the number of training parameters.}

    \label{table:parameters}
    \begin{tabular}{lcl}
      \textit{Model} & \textit{Trained layers} & \textit{Network param}\\
      \hline
      Source  & all (11 layers) & 470466\\
      \hline
      Target 3 layers  & FC1 + FC2 + FC3 & 172928\\
      Target 2 layers  & FC2 + FC3 & 41344\\
      Target 1 layer  & FC3 & 8320\\
      \hline
    \end{tabular}
  \end{center}
\end{table}

\section{Experiments}
\label{sec:experiments}

\subsection{Clinical MS dataset}
\label{subsec:label}

\subsubsection{Data}
A total of 60 patients with a clinically isolated syndrome (Hospital Vall d'Hebron, Barcelona, Spain) were scanned on a 3T Siemens with a 12-channel phased-array head coil (Trio Tim, Siemens, Germany) with the following acquired sequences: 1) transverse DP/T2-w fast spin-echo (TR=2500 ms, TE=16-91 ms, voxel size=0.78$\times$0.78$\times$3 mm$^3$),  2) transverse fast T2-FLAIR (TR=9000 ms, TE=93 ms, TI=2500 ms, flip angle=120$^{\circ}$, voxel size=0.49$\times$0.49$\times$3 mm$^3$), and 3) sagittal 3D T1 magnetization prepared rapid gradient-echo (MPRAGE) (TR=2300 ms, TE=2 ms, TI=900 ms, flip angle=9$^{\circ}$; voxel size=1$\times$1$\times$1.2 mm$^3$). For each patient, WM lesion masks were semi-automatically delineated from either PD or FLAIR masks using JIM software\footnote{\texttt{Xinapse Systems, http://www.xinapse.com/home.php}} by an expert radiologist of the same hospital center. The T1-w and FLAIR images were first skull-stripped using BET \citep{Smith2002} and intensity normalized using N3 \citep{Sled1998}. The FLAIR images were affinely co-registered to the T1-w space using the FSL-FLIRT utility \citep{Greve2009}.

\subsubsection{Evaluation:}

Th images were first randomly split into two sets composed of 30 training and testing images. Then, the training data were used to train the different target models while accounting for the following factors:
\begin{itemize}
\item The effect of one-shot domain adaptation training. Each proposed domain adaptation configuration was trained using a single training image with a lesion size in the range of [0.5-18] ml.
\item The effect of the proposed domain adaptation configurations on the accuracy of the target model (retraining 1, 2 or all of the FC layers, see Table \ref{table:parameters}).
\item The effect of the number of training images used to re-train the target model. Each proposed domain adaptation configuration was trained using 1, 2, 5, 10, 15 or all of the available training images.
\end{itemize}

After training, each of the target models was feed-forwarded on the test set, evaluating the accuracy of the resulting segmentations against the available lesion annotations using the following evaluation metrics:

\begin{itemize}

\item The overall \% segmentation accuracy in terms of the dice similarity coefficient ($DSC$) between the manual lesion annotations and the output segmentation masks:
  \begin{equation}
    DSC = \frac{2\times{TP_s}}{FN_s+FP_s+2\times{TP_s}} \times 100
  \label{eq:dsc}
\end{equation}
\noindent where $TP_s$ and $FP_s$ denote the number of voxels correctly and incorrectly classified as a lesion, respectively, and $FN$ denotes the number of voxels incorrectly classified as a non-lesion.

\item Sensitivity of the method in detecting lesions between manual lesion annotations and output segmentation masks, expressed in \%:
  \begin{equation}
sensitivity = \frac{TP_d}{TP_d+FN_d} \times 100
  \label{eq:tp}
\end{equation}

\noindent where $TP_d$ and $FN_d$ denote the number of correctly and missed lesion region candidates, respectively.

\item Precision of the method in detecting lesions between manual lesion annotations and output segmentation masks, also expressed in \%:
  \begin{equation}
  precision = \frac{TP_d}{TP_d+FP_d} \times 100
  \label{eq:fp}
\end{equation}

\noindent where $TP_d$ and $FP_d$ denote the number of correctly and incorrectly classified lesion region candidates, respectively.

\end{itemize}

To evaluate the effectiveness of the proposed framework, the obtained results were compared against the source model without re-training and the same target model fully trained using all of the available training images. For comparison, the segmentation accuracies of two state-of-the-art MS lesion segmentation pipelines LST \citep{Schmidt2012} and SLS \citep{Roura2015}, were also reported.

\subsubsection{Experiment details}

All of the training images were first normalized with a zero mean and standard deviation of one. Each of the trained models was run with the exact parameters used to train the source model (see Subsection \ref{subsec:source_details}). The number of lesion voxels was equal during all of the training epochs. Normal appearing tissue voxels were re-sampled every 10 epochs to augment the tissue variability during the training. As in the source model, the post-processing parameters $\ge t_{bin}$ and $l_{min}$ were set to 0.5 and 10, respectively. In the LST, the parameters $\kappa$ and $lgm$ were optimized for the current dataset with the values $\kappa=0.15$ and $lgm=gm$, respectively. In the SLS, the parameters $\alpha, \lambda_{ts}$ and $ \lambda_{ns}$ were also optimized for this particular dataset with the values $\alpha=3$, $\lambda_{ts}=0.6$ and $\lambda_{nb}=0.6$ for both iterations.

\begin{table}[tp]
  \vspace{-0.5cm}
  \scriptsize
  \caption{Clinical MS dataset: DSC, sensitivity and precision coefficients for each of the models re-trained using a single image with varying degree of lesion load. For comparison, the obtained values for SLS \citep{Roura2015}, LST \citep{Schmidt2012} and the same cascaded CNN method fully trained using the entire training dataset \citep{Valverde2017} are also shown. For each coefficient, the reported values are the mean (standard deviation) when evaluated on the 30 testing images.}
  \label{table:oneshot_vh}
  \begin{center}
    \begin{tabular}{llll}
      lesion vol (num lesions) & DSC & sensitivity & precision\\
      \hline
      \multicolumn{4}{c}{1 layer (FC3)} \\
      \hline
      0.5 ml (9 lesions) & 0.30 (0.19) & 0.44 (0.23) & 0.49 (0.30)\\
      1.2 ml (11 lesions) & 0.39 (0.19) & 0.44 (0.19) & 0.67 (0.23)\\
      3.1 ml (17 lesions) & 0.38 (0.22) & 0.46 (0.20) & 0.54 (0.25)\\
      8.3 ml (90 lesions) & 0.44 (0.17) & 0.58 (0.19) & 0.58 (0.26)\\
      18  ml (78 lesions) & 0.47 (0.18) & 0.59 (0.18) & 0.58 (0.23)\\
      \hline
      \multicolumn{4}{c}{2 layers (FC2 + FC3)} \\
      \hline
      0.5 ml (9 lesions)  & 0.30 (0.17) & 0.52 (0.23) & 0.54 (0.28)\\
      1.2 ml (11 lesions) & 0.39 (0.18) & 0.49 (0.21) & 0.72 (0.29)\\
      3.1 ml (17 lesions) & 0.36 (0.22) & 0.42 (0.20) & 0.54 (0.27)\\
      8.3 ml (90 lesions) & 0.45 (0.15) & 0.55 (0.18) & 0.66 (0.24)\\
      18  ml (78 lesions) & 0.44 (0.19) & 0.62 (0.20) & 0.52 (0.25)\\
      \hline
      \multicolumn{4}{c}{3 layers (FC1 + FC2 + FC3)} \\
      \hline
      0.5 ml (9 lesions)   & 0.28 (0.17) & 0.48 (0.22) & 0.48 (0.28)\\
      1.2 ml (11 lesions)  & 0.38 (0.17) & 0.52 (0.22) & 0.72 (0.26)\\
      3.1 ml (17 lesions)  & 0.38 (0.21) & 0.46 (0.21) & 0.55 (0.25)\\
      8.3 ml (90 lesions)  & 0.44 (0.17) & 0.61 (0.17) & 0.57 (0.26)\\
      18  ml (78 lesions)  & 0.45 (0.18) & 0.60 (0.21) & 0.55 (0.23)\\
      \hline
      \textit{Source} (0 lesions) & 0.23 (0.22) & 0.42 (0.43) & 0.45 (0.34) \\
      SLS & 0.25 (0.17) & 0.34 (0.25) & 0.51 (0.30)\\
      LST & 0.28 (0.23) & 0.31 (0.21) & 0.59 (0.27)\\
      CNN & 0.53 (0.16) & 0.60 (0.21) & 0.75 (0.21)\\
      \hline
    \end{tabular}
  \end{center}
\end{table}

\subsubsection{Results}

First, we evaluated the models under a one-shot domain adaptation scenario, by training them again several times using only a single image from the training set with lesion burdens equal to 0.5, 1.2, 3.1, 8.3 and 18 ml. Table \ref{table:oneshot_vh} shows the DSC, sensitivity and precision coefficients of each of the re-trained models under
different one-shot training sets. The same evaluation is also shown for LST, SLS, and the cascaded CNN architecture without fine-tunning (\textit{source}) and fully trained using the entire training dataset. As expected, the model without domain adaptation reported the worst accuracy by the lack of adaptability of the source knowledge. In contrast, the models performance increased with the number of annotated lesions on the target domain, showing better overlap with the manual annotations than LST and SLS, even in extreme cases in which only 9 lesions are manually annotated on the target domain (0.5 ml).

\begin{figure*}[tp]
  \begin{center}
      \includegraphics[width=1.0\textwidth]{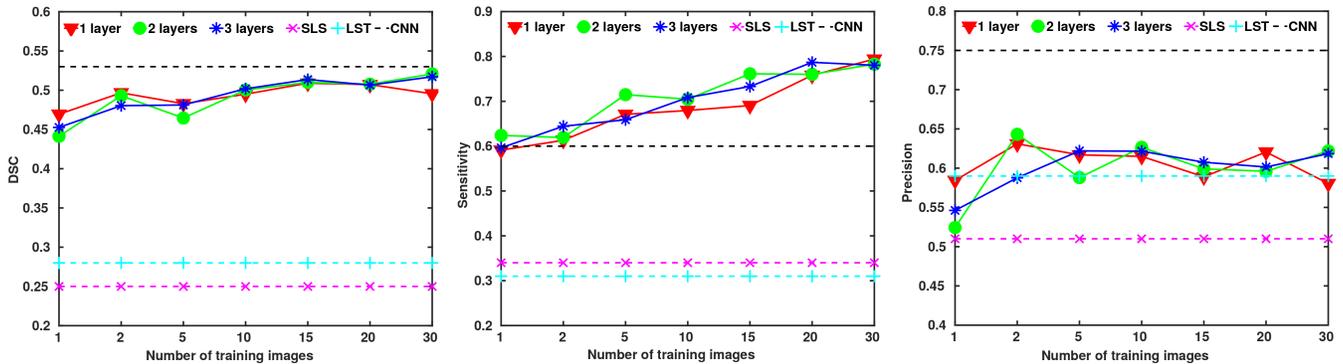}
  \end{center}
  \caption{Effect of the number of re-trained FC layers and training images on the DSC, sensitivity and precision coefficients when evaluated on the clinical MS dataset. The represented value for each configuration is computed as the mean DSC, sensitivity and precision scores over the 30 testing images. For comparison, the obtained values for the lesion segmentation methods SLS \citep{Roura2015} (  $\times$ pink line), LST \citep{Schmidt2012} ($+$ cyan line) and the same cascaded CNN method fully trained using all of the available training data \citep{Valverde2017} (- black line) are shown.}
  \label{fig:clinical_experiment_all}
\end{figure*}

As a second experiment, we evaluated the effect of adding more training data on the accuracy of the domain adapted models. Figure \ref{fig:clinical_experiment_all} shows the DSC, sensitivity and precision coefficients of each of the re-trained models using different number of training image patients which ranged from 1 to 30. The number of training samples was $\sim18K, \sim36k,\sim48k,\sim60K,\sim70K,\sim95K$ and $\sim130K$ for 1, 2, 5, 10, 15, 20 and 30 images, respectively. When more training data on the target space were available, the performances of the re-trained models were similar to that of the fully trained CNN pipeline, especially those of the models in which the last two or all of the FC layers were re-trained. In contrast, in the sensitivity and precision plots, the re-trained models were in general more sensitive to inferring WM lesions but at the cost of increasing also the number of false-positive outcomes.

\subsection{ISBI 2015 dataset}
\label{subsec:label}
\subsubsection{Data}
The ISBI2015 MS lesion challenge \citep{Carass2017} was composed of 5 training and 14 testing subjects with 4 or 5 different image time-points per subject.  All of the data were acquired on a 3.0 Tesla MRI scanner (Philips Medical Systems, Best, The Netherlands) with  T1-w MPRAGE, T2-w, PD and FLAIR sequences. A complete description of the image protocol and pre-processing details is available on the organizer's website \footnote{\texttt{http://iacl.ece.jhu.edu/index.php/MSChallenge/data}}. On the challenge competition, each subject image was evaluated independently, which led to a final training and testing sets composed of 21 and 61 images, respectively. Additionally, manual delineations of MS lesions performed by two experts were included for each of the 21 training images.

The evaluation of the ISBI 2015 challenge is performed blind for the teams by submitting the segmentation masks of the 61 testing cases to the challenge website evaluation platform\footnote{\texttt{https://smart-stats-tools.org/node/26}}. Different metrics are computed as part of an overall performance score \citep{Carass2017}, where values above 90 are considered to be comparable to human performance.

\subsubsection{Evaluation}
Here, we analyzed the effect of one-shot domain adaptation on the overall performance of the testing set. To do so, we retrained all of the model configurations (1, 2 or all FC layers) with a single training image from each training subject, which led to 5 different training sets with varying number of lesions and a total lesion volume in the range [2.3-26.8 ml]. Then, each of the resulting trained models was
feed-forwarded on the blind test set. Based on that approach, we evaluated the following experiments:

\begin{itemize}
\item The effect of the number lesions and lesion volume on the performance of each of the one-shot domain adaptation models. We considered the segmentation masks of the same cascaded architecture fully trained using the 21 training images \citep{Valverde2017} as silver mask annotations, given that this particular model already reported human-like accuracy (score 91.44) when submitted to the challenge platform (4th position / 46 participants). We evaluated the performance of each of the one-shot models again while computing the DSC, sensitivity and precision coefficients between the one-shot segmentation masks and the silver masks.

\item The performance of the best one-shot domain adaptation model on the blind test set. The best performing model from the previous experiment was sent to the challenge's evaluation platform, comparing its accuracy to those of the other submitted MS lesion segmentation pipelines fully trained using the entire available training set. Among the set of evaluated coefficients computed in the challenge, only the DSC, sensitivity and precision metrics are shown for comparison.

\end{itemize}

\subsubsection{Experimental details}
Like in the clinical MS dataset, all of the training images were first normalized with a zero mean and a standard deviation of one. Each of the trained models was run with the exact parameters used to train the source model (see Subsection \ref{subsec:source_details}). The number of lesion voxels was equal during all of the training epochs. Normal appearing tissue voxels were re-sampled every 10 epochs to augment the tissue variability during the training. The post-processing parameters $\ge t_{bin}$ and $l_{min}$ were set also to 0.5 and 10, respectively.

\subsubsection{Results}

\begin{table}[tp]
  \scriptsize
  \caption{ISBI dataset:  DSC, sensitivity and precision coefficients for each of the models re-trained using a single image of the training dataset against the silver masks. For comparison, the obtained values for the same \textit{source} CNN method without domain adaptation (see Subsection \ref{subsec:initial_training}) are also shown. For each coefficient, the reported values are the mean (standard deviation) when evaluated on the 61 testing images.}
  \label{table:isbi_oneshot}
  \begin{center}
    \begin{tabular}{llll}
      lesion vol (num lesions) & DSC & sensitivity & precision\\
      \hline
      \multicolumn{4}{c}{1 layer (FC3)} \\
      \hline
      ISBI01 (17.4 ml, 29 lesions) & 0.56 (0.14) & 0.80 (0.11) & 0.62 (0.07)\\
      ISBI02 (26.8 ml, 45 lesions) & 0.51 (0.21) & 0.83 (0.13) & 0.55 (0.07)\\
      ISBI03 (5.9 ml, 26 lesions) & 0.65 (0.11) & 0.60 (0.17) & 0.80 (0.14))\\
      ISBI04 (2.3 ml, 20 lesions) & 0.33 (0.12) & 0.41 (0.16) & 0.81 (0.14)\\
      ISBI05 (4.3 ml, 22 lesions) & 0.54 (0.11) & 0.56 (0.16) & 0.84 (0.12)\\
      \hline
      \multicolumn{4}{c}{2 layers (FC2 + FC3)} \\
      \hline
      ISBI01 (17.4 ml, 29 lesions) & 0.56 (0.14) & 0.74 (0.11) & 0.59 (0.06)\\
      ISBI02 (26.8 ml, 45 lesions) & 0.53 (0.21) & 0.87 (0.11) & 0.56 (0.06)\\
      ISBI03 (5.9 ml,  26 lesions) & 0.65 (0.11) & 0.66 (0.15) & 0.79 (0.13)\\
      ISBI04 (2.3 ml,  20 lesions) & 0.47 (0.12) & 0.48 (0.18) & 0.83 (0.11)\\
      ISBI05 (4.3 ml,  22 lesions) & 0.56 (0.11) & 0.54 (0.16) & 0.82 (0.13)\\
      \hline
      \multicolumn{4}{c}{3 layers (FC1 + FC2 + FC3)} \\
      \hline
      ISBI01 (17.4 ml ,29 lesions) & 0.66 (0.10) & 0.73 (0.11 & 0.78 (0.10)\\
      ISBI02 (26.8 ml ,45 lesions) & 0.69 (0.13) & 0.70 (0.18) & 0.77 (0.10)\\
      ISBI03 (5.9 ml, 26 lesions) & 0.65 (0.11) & 0.63 (0.13) & 0.79 (0.14)\\
      ISBI04 (2.3 ml, 20 lesions) & 0.47 (0.14) & 0.40 (0.16) & 0.84 (0.08)\\
      ISBI05 (4.3 ml, 22 lesions) & 0.46 (0.12) & 0.46 (0.17) & 0.87 (0.13)\\
      \hline
      \textit{Source} (0 lesions) & 0.33 (0.12) & 0.40 (0.16) & 0.72 (0.14)\\
      \hline
    \end{tabular}
  \end{center}
\end{table}

\begin{figure*}[tp]
  \vspace{-1cm}
  \begin{center}
      \includegraphics[width=0.9\textwidth]{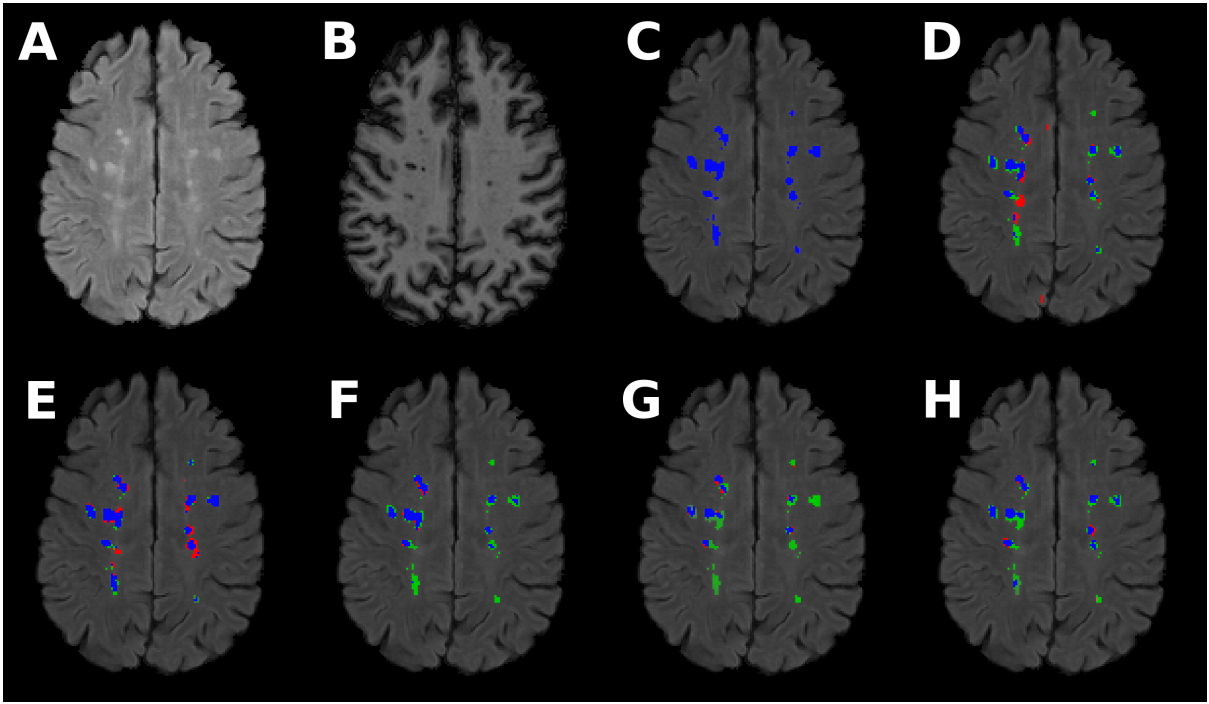}
  \end{center}
  \caption{Output segmentation masks for the first image of the ISBI testing set. (A) FLAIR and (B) T1-w input masks. Silver mask (C) obtained based on the same CNN method fully trained on the entire training dataset \citep{Valverde2017}. The other panels show the output masks for the one-shot domain adaptation model re-trained only for the last FC layer using the images (D) ISBI01 (17.4 ml), (E) ISBI02 (26.8 ml), (F) ISBI03 (5.9 ml), (G) ISBI04 (2.3 ml), and (H) ISBI05 (4.3 ml).  The blue regions depict the overlapped lesions between the silver mask and each of the models. The red and green regions depict false-positive and false-negative lesions, respectively, with respect to the silver mask.}
    \label{fig:isbi_qualitative}
\end{figure*}

Table \ref{table:isbi_oneshot} shows the performance of each of the one-shot domain adaptation models when trained on different images with varying degrees of lesion size. For comparison, the results for the source model without re-training on the target domain are also depicted. The performance of the source model pre-trained only on the MICCAI2008 and MICCAI2016 datasets shows the lack of accuracy of the method in delineating WM lesions on the unseen target domain. Following the same pattern seen on the clinical MS dataset, the best performance with respect to the silver masks was obtained when re-training all of the FC layers with the maximum number of available voxels (ISBI02, 26.8 ml.). Interestingly, the performance of the model re-trained using just 26 lesions (ISBI03, 5.9 ml.) was remarkably higher than that of the other trained models, especially when only the last two or one FC layers were re-trained. Figure \ref{fig:isbi_qualitative} depicts the effect of the available number of lesion voxels on the resulting number of true-positive, false-positive and false-negative outcomes when re-training only the last FC layer.

Table \ref{table:isbi_challenge} depicts the performance of the best domain adaptation model (ISBI02 with 3 re-trained layers) against different top rank participant challenge strategies. From the list of compared methods, the best five strategies were based on CNN models  \citep{Andermatt2017a, Hashemi2018, Valverde2017, Greenspan2016, Roy2018}, while the others were based on either other supervised learning techniques \citep{Valcarcel2018, Deshpande2015, Sudre2015} or unsupervised intensity models \citep{Shiee2010, Jain2015}. The accuracy of the one-shot domain model was similar to those of other recently fully trained submitted CNN models \citep{Roy2018}, yielding a performance that was comparable to human performance (score 90.3), even when trained it with a single training image. Furthermore, the proposed one-shot method reported a performance similar to that of the same fully trained cascaded CNN architecture (score 91.44) \citep{Valverde2017}, which shows the capability of the model to adapt the source knowledge into the target domain using a reduced training dataset.

\section{Discussion}

Several CNN methods have been proposed for automated MS lesion segmentation, in most of the cases showing a performance similar to that of human expert raters. However, the performance of these models tend to decrease significantly when evaluated on image domains other than those used for training the model, thus showing a lack of adaptability to unseen data. In this paper, we have studied the effect of intensity domain adaptation on our recently published CNN-based MS lesion segmentation method. The model was fully trained on two public MS lesion datasets (MICCAI2008, MICCAI2016), analyzing its capability to transfer the acquired knowledge to two completely unrelated datasets. For this particular architecture, we evaluated the number of necessary layers that must be retrained and the minimum number of annotated images from the unseen domain that is required to obtain a similar fully trained performance.

Although the small number of network parameters of our cascaded architecture used as a source model ($\sim470K$), a considerable number of training images was still required to optimize the entire set of parameters. In this regard, our experiments on the clinical MS dataset show that when using the whole set of available training images, the performances of the models in which only the FC layers were re-trained were very similar to that of the same model fully trained for both the convolutional and FC layers. This result suggests that there is an inherent capability of the convolutional layers to encode useful image features that can be used across different image domains without re-adaptation. As shown in Table \ref{table:parameters}, by re-using some of the network layers we drastically reduce the number of parameters to optimize on the target domain, and thus, the domain-adapted networks can be fitted using a small number of training samples without over-fitting the model.

\begin{table*}[tp]
  \scriptsize
  \caption{ISBI challenge: DSC, sensitivity, precision and overall score coefficients for the best one-shot domain adaptation model (ISBI02 with 3 layers) after submitting the segmentation masks for blind evaluation. The obtained results are compared with different top rank participant strategies. For each method, the reported values are extracted from the challenge results board. The reported values are the mean (standard deviation) when evaluated on the 61 testing images. The performance of the methods with an overall $score\geq 90$ is considered to be similar to human performance.}
  \label{table:isbi_challenge}
  \begin{center}
    \begin{tabular}{llllr}
      Method & DSC & sensitivity & precision & score\\
      \hline
      \citet{Andermatt2017a} & 0.63 (0.14) & 0.54 (0.19) & 0.84  (0.10) & 92.07\\
      \citet{Hashemi2018} & 0.66 (0.11) & 0.67 (0.20) & 0.71  (0.16) & 91.52\\
      \citet{Valverde2017} & 0.64 (0.12) & 0.57 (0.17) & 0.79 (0.15) & 91.44\\
      \citet{Greenspan2016} & 0.63 (0.14) & 0.55 (0.18) & 0.80  (0.15) & 91.26\\
      \citet{Roy2018}* & 0.52 (-~-) & -~- (-~-) & 0.86 (-~-) & 90.48\\
      \citet{Deshpande2015} & 0.60 (0.13) & 0.55 (0.17) & 0.73  (0.18) & 89.81\\
      \citet{Jain2015} & 0.55 (0.14) & 0.47 (0.15) & 0.73  (0.20) & 88.74\\
      \citet{Shiee2010} & 0.55 (0.19) & 0.54 (0.15) & 0.70  (0.29) & 88.46\\
      \citet{Valcarcel2018} & 0.57 (0.13) & 0.57 (0.18) & 0.61  (0.16) & 87.71\\
      \citet{Sudre2015} & 0.52 (0.14) & 0.46 (0.15) & 0.66  (0.18) & 86.44\\
      \hline
      one-shot (3 layers, 26.8 ml.) & 0.58 (0.16) & 0.48 (0.19) & 0.84 (0.13) & 90.32\\
      \hline
    \end{tabular}
    \\~\\(*) Obtained results for \citet{Roy2018} were extracted from the related publication.
  \end{center}
\end{table*}

Our experiments highlight the relationship between the number of available lesion samples used to re-train the model and the resulting accuracy.  As seen in the first experiment, the incorporation of additional training samples increase the segmentation overlap (DSC) on all of re-trained models. As expected, the adaptation of two or all FC layers was progressively more effective than that of adapting only the last FC layer when increasing the lesion samples, since the additional characteristics of the target dataset could be fine-tuned on the FC1 and FC2 layers. The sensitivity and precision of all of the domain-adapted methods also increased remarkably with the training data. The addition of progressively more lesion and normal appearing patches increased the confidence of all of the adapted models, thus reducing the number of false-positive lesion voxels.

More interestingly, the models still yielded a remarkably high performance on reduced training sets, such as a single training image. On the clinical MS dataset, the performances of the one-shot adapted models were significantly higher than those of the LST and SLS, even when trained using a single image with a 3.1 ml. lesion load and 17 manual annotated regions. Although the SLS and LST methods were unsupervised models that did not require strict training, their parameters were optimized for the target image domain using a time consuming grid-search. In the ISBI2015 challenge, the same cascaded CNN model fully trained on the 21 training images performed in the top rank (4th position / 46 participants), yielding comparable human-like accuracy. When compared with this fully trained model, the accuracy of the one-shot domain-adapted model trained with only one of the 21 training images was still remarkably higher than those of most of the participant strategies, which was very similar to other CNN methods and still yielded a comparable human-like accuracy. This finding  is relevant, and it shows the potential applicability of our cascaded CNN method on very reduced datasets with a limited loss in the accuracy.

In general, none of the hyper-parameters optimized for the source model were fine-tuned on any of the domain-adapted models, which kept them fixed along of all the experiments conducted in this study. As previously observed, for a training dataset that contained at least 3000 lesion voxels (3 ml. on a isotropic $1 mm^3$), the best results were obtained when the last two or all of the FC layers were re-adapted.  In contrast, on extremely small datasets of $<3$ ml., re-training only the last layer appeared to be more indicative in order reducing the over-fitting of the model.  Given that these parameters appeared to work well in most of the datasets, we propose using them as a rule of thumb on future settings.

\section{Conclusions}
\label{sec:conclusion}
In this study, we analyzed the effect of intensity domain adaptation on a recent CNN-based MS lesion segmentation method. Given a source model trained on two public MS datasets, we studied how transferable the acquired knowledge was when applied to a private dataset and the ISBI2015 challenge dataset, upon evaluating the minimum number of annotated images needed from the new domain and the minimum number of layers needed to re-train to obtain a comparable accuracy.

Our experiments showed the effectiveness of the proposed domain adaptation model in transferring previously acquired knowledge to new image domains even if only a single training image was available on the target dataset. On the ISBI2015, the accuracy of our one-shot domain-adapted model was comparable to that of a human expert rater and similar to those of other CNN methods trained on a wide set of training data. In this aspect, we believe that the performance shown by our domain adapted models will encourage the MS community to incorporate its use in different clinical settings with reduced amounts of annotated data. This finding could be meaningful not only in terms of the accuracy in delineating MS lesions but also in the related reductions in time and economic costs derived from manual lesion labeling.

\section*{Acknowledgements}
Mariano Cabezas holds a Juan de la Cierva - Incorporaci\'on grant from the Spanish Government with reference number IJCI-2016-29240. This work has been partially supported by La Fundaci\'o la Marat\'o de TV3, by Retos de Investigaci\'on TIN2014-55710-R, TIN2015-73563-JIN and DPI2017-86696-R from the Ministerio de Ciencia y Tecnolog\'ia. The authors gratefully acknowledge the support of the NVIDIA Corporation with their donation of the TITAN-X PASCAL GPU used in this research.

\bibliographystyle{apalike}
\bibliography{bibtex}

\end{document}